\pdfoutput=1

\documentclass[11pt]{article}

\usepackage[final]{acl}

\usepackage{times}
\usepackage{latexsym}

\usepackage[T1]{fontenc}

\usepackage[utf8]{inputenc}

\usepackage{microtype}

\usepackage{inconsolata}

\usepackage{graphicx}

\usepackage{multirow}
\usepackage{enumitem}
\usepackage{booktabs}
\usepackage{array}
\usepackage{makecell}
\usepackage{todonotes}
\usepackage{xcolor}
\usepackage{tabularx}

\title{
  MessIRve: A Large-Scale Spanish Information Retrieval Dataset
}

\author{
 \textbf{Francisco Valentini\textsuperscript{1}},
 \textbf{Viviana Cotik\textsuperscript{2,1}},
 \textbf{Damián Furman\textsuperscript{1}},
 \textbf{Ivan Bercovich\textsuperscript{3}},
\\
 \textbf{Edgar Altszyler\textsuperscript{4}},
 \textbf{Juan Manuel Pérez\textsuperscript{1}}
\\
\\
 \textsuperscript{1}CONICET-Universidad de Buenos Aires.\\Instituto de Ciencias de la Computación (ICC). Buenos Aires, Argentina\\
 \textsuperscript{2}Universidad de Buenos Aires. Facultad de Ciencias Exactas y Naturales.\\Departamento de Computación. Buenos Aires, Argentina.\\
 \textsuperscript{3}University of California, Santa Barbara. California, U.S.A.\\
 \textsuperscript{4}Quantit\\
   \small{\texttt{\{fvalentini,jmperez,vcotik\}@dc.uba.ar},}
   \small{\texttt{damian.a.furman@gmail.com},}\\[-0.3em]
   \small{\texttt{ibercovich@ucsb.edu},}
   \small{\texttt{edgar@bequantit.com}}
}

\begin{document}
\maketitle

\begin{abstract}
  Information retrieval (IR) is the task of finding relevant documents in response to a user query. Although Spanish is the second most spoken native language, there are few Spanish IR datasets, which limits the development of information access tools for Spanish speakers. We introduce MessIRve, a large-scale Spanish IR dataset with almost 700,000 queries from Google's autocomplete API and relevant documents sourced from Wikipedia. MessIRve's queries reflect diverse Spanish-speaking regions, unlike other datasets that are translated from English or do not consider dialectal variations. The large size of the dataset allows it to cover a wide variety of topics, unlike smaller datasets. We provide a comprehensive description of the dataset, comparisons with existing datasets, and baseline evaluations of prominent IR models. Our contributions aim to advance Spanish IR research and improve information access for Spanish speakers.
\end{abstract}

\section{Introduction} \label{sec:introduction}

Given a user query, Information Retrieval (IR) is the process of finding and ranking relevant documents from a collection of data. IR systems have become increasingly important as they are the backbone for the currently widespread Retrieval Augmented Generation (RAG) systems, in which IR provides relevant passages that are subsequently fed into a Large Language Model (LLM, \citealp{lewis2020retrieval,izacard2023atlas,ram2023incontext,shi2024replug}). This approach has been shown to improve the quality of generated text, reducing the likelihood of factual errors or hallucinations by LLMs in several tasks \citep{cheng2023uprise,jiang2023active,cheng2024lift,lin2024radit}.

Natural language processing relies heavily on evaluation datasets, which establish a common ground for comparing algorithms and guide the development of new methods. While a wealth of resources are available for English, other languages often rely on multilingual models and datasets.

When addressing tasks in Spanish, resources are notably scarce. Spanish is the the second most spoken language in the world by number of native speakers, with over 486 million speakers. It is also the official language of 20 countries and is widely spoken in the United States \citep{eberhard2024ethnologue}. Despite its prevalence, at the time of writing this paper, the retrieval section of the popular MTEB benchmark \citep{muennighoff-etal-2023-mteb} includes datasets for English, Chinese, French, and Polish, but omits Spanish.\footnote{\href{https://huggingface.co/spaces/mteb/leaderboard}{huggingface.co/spaces/mteb/leaderboard}}

The scarcity of evaluation datasets is a major challenge for the development of IR systems for Spanish. 
Existing Spanish IR datasets are limited and present shortcomings in terms of dataset size, reliance on translations, restricted topical coverage, or lack of dialectal diversity.
Which systems are best for Spanish IR? 
This key question remains unanswered due to the lack of a comprehensive Spanish IR benchmark.

In this work, we address this gap by:

\begin{enumerate}[wide, itemindent=\labelsep, itemsep=0pt]

  \item Identifying and describing the Spanish IR datasets that are currently available.

  \item Introducing MessIRve, a new large-scale dataset in Spanish that addresses current datasets' limitations through automated data collection, the use of natural, non-translated queries, and the inclusion of dialectal diversity across the Spanish-speaking countries.\footnote{MessIRve means \emph{works for me} in Spanish (``me sirve''). The reference to Messi, player of the most popular sport in Spanish-speaking countries, football, also stresses the importance of using topics that are relevant to Spanish speakers.}

  \item Providing baseline evaluations of relevant models on the new dataset and the existing ones.

  \item Making the dataset and a fine-tuned baseline model publicly available, so that other researchers can use them to develop and evaluate their own IR systems for Spanish.\footnote{
      Dataset: \href{https://huggingface.co/datasets/spanish-ir/messirve}{hf.co/datasets/spanish-ir/messirve} \\
      Model: \href{https://huggingface.co/spanish-ir/multilingual-e5-large-ft-messirve-full}{hf.co/spanish-ir/multilingual-e5-large-ft-messirve-full} \\
      Code: \href{https://github.com/ftvalentini/MessirveSpanishIR}{github.com/ftvalentini/MessirveSpanishIR}
  }

\end{enumerate}

These contributions aim to spur further research in IR for the Spanish language and to facilitate the development of efficient information access tools for Spanish speakers.

\section{Related work} \label{sec:related_work}

\begin{table*}[ht!]
  \centering
  \begin{tabular}{ l c cc cc cc }
    \toprule
    \multirow{2.5}{*}{\textbf{Dataset}} & \multirow{2.5}{*}{\textbf{\#Docs}} & \multicolumn{2}{c}{\textbf{Train}} & \multicolumn{2}{c}{\textbf{Dev}} & \multicolumn{2}{c}{\textbf{Test}}                                                           \\
    \cmidrule(lr){3-4} \cmidrule(lr){5-6} \cmidrule(lr){7-8}
                                        &                                     & \textbf{\#Q}                      & \textbf{\#J/\#Q} & \textbf{\#Q} & \textbf{\#J/\#Q} & \textbf{\#Q} & \textbf{\#J/\#Q} \\
    \midrule\midrule
    \thead[l]{mMARCO     \\ \citep{bonifacio2021mmarco}}      & 8 841 823   & 532 761     & 1.00      & -        & -      & 6 979    & 1.07       \\
    \thead[l]{SQAC       \\ \citep{gutierrez2022maria}}       & 6 247       & 14 934      & 1.01      & 1 861    & 1.00   & 1 908    & 1.00       \\
    \thead[l]{MIRACL     \\ \citep{zhang2023miracl}}          & 10 373 953  & 2 162       & 9.96      & 648      & 9.94   & 1 515    & 9.95       \\
    \thead[l]{PRES       \\ \citep{kamateri2019test}}         & 10 037      & -           & -         & -        & -      & 167      & 7.72       \\
    \thead[l]{Multi-EuP  \\ \citep{yang-etal-2023-multi-eup}} & 2 371       & -           & -         & -        & -      & 633      & 3.75       \\
    \midrule
    \textbf{MessIRve (\texttt{full})}                                      & 14 047 759  & 537 730     & 1.06         & -        & -      & 156 528  & 1.02    \\
    \bottomrule
\end{tabular}

  \caption{\textbf{Spanish IR Datasets.}
    \#: number of, Docs: documents, Q: queries, J: annotated relevance judgments. Judgments indicate positive (relevant) documents in all datasets, except for MIRACL, which includes negative (non-relevant) annotations as well. The \texttt{full} split of MessIRve includes queries that are not specific to any country plus the set of unique queries from all countries (see Section \ref{sec:data_splits}).
  }
  \label{tab:current_datasets}
\end{table*}

In this section, we review well-documented, publicly available Spanish IR datasets. Table \ref{tab:current_datasets} presents their statistics. A dataset consists of queries, documents, and relevance judgments \citep{thakur2021beir}. Queries express information needs, documents contain potential answers, and relevance judgments assess document-query relevance. The full document set is the corpus.

The multilingual MSMARCO (\textbf{mMARCO}, \citealp{bonifacio2021mmarco}) translates the well-known MSMARCO dataset \citep{bajaj2016ms} into Spanish using machine translation. Though not originally in Spanish, it provides a large-scale resource for training IR models, leveraging Bing search queries. Annotators judged the relevance of 10 Bing-retrieved passages per query, which form the document collection.

The Spanish Question Answering Corpus (\textbf{SQAC}, \citealp{gutierrez2022maria}) is a non-translated dataset inspired by SQUAD \citep{rajpurkar-etal-2016-squad}. Spanish-speaking annotators generated up to five questions per passage from Spanish Wikipedia and the AnCora corpus \citep{taule2008ancora}, encouraging synonym use. While lacking a full document collection, the set of relevant passages can serve as an IR corpus.

\textbf{MIRACL} (Multilingual Information Retrieval Across a Continuum of Languages, \citealp{zhang2023miracl}) includes Spanish and 17 other languages, emphasizing high-quality native speaker relevance judgments. Some annotators created queries from Wikipedia paragraphs, and others assessed the relevance of 10 passages retrieved by different models. Despite containing relatively few queries, MIRACL benefits from having more annotations per query and also negative relevance judgments.

Beyond multi-domain datasets, two domain-specific ones exist. The Spanish Passage Retrieval dataset (\textbf{PRES}, \citealp{kamateri2019test}) has health-related queries and articles from reliable Spanish health websites as documents. Relevance judgments were done via pooling, with annotators reviewing top-ranked results by different IR systems.

Finally, the Multilingual European Parliament Dataset (\textbf{Multi-EuP}, \citealp{yang-etal-2023-multi-eup}) uses European Parliament debate topics as queries, with speeches by Spanish parliamentarians as documents. Relevance judgments indicate whether a speech was part of a debate on a given topic.

\section{MessIRve} \label{sec:dataset}

MessIRve is a new large-scale dataset for Spanish IR, designed to better capture the information needs of Spanish speakers across different countries. In this section we describe the data collection process (Section \ref{sec:data_collection}) and the dataset splits (Section \ref{sec:data_splits}), and perform a quality assessment (Section \ref{sec:quality_assessment}).

\subsection{Data collection} \label{sec:data_collection}

MessIRve is built by using questions from Google's autocomplete API\footnote{Endpoint: \url{www.google.com/complete/search?}} as queries, and answers from Google Search ``featured snippets'' that link to Wikipedia as relevant documents. This data collection strategy is inspired by the GooAQ dataset \citep{khashabi-etal-2021-gooaq-open}.

\paragraph{Queries} Starting with 120 predefined prefixes such as ``qué'' (\emph{what}), ``cómo'' (\emph{how}), ``dónde'' (\emph{where}), ``quién'' (\emph{who}), etc., we used the Google autocomplete API to obtain popular queries in Spanish starting with these prefixes (refer to Appendix \ref{app:queries}, Table \ref{tab:google_prefixes} for the list of prefixes). We then iteratively expanded the set of prefixes based on the returned results until we approximately reached a predefined number of results. The API does not always return queries that begin strictly with the provided prefixes, resulting in a dataset where not all queries begin with the initial prefixes (see further details on query extraction in Appendix \ref{app:queries}).

We ran this process for 20 countries with Spanish as an official language, plus the United States, by adjusting the API parameters to return results in Spanish for each country. Equatorial Guinea was the only country left out because it doesn't have a Google domain. The extraction took place between March and April 2024, so the queries may reflect popular interests during that period.

During the process we noticed that some API results were independent of the country-specific domain, so we obtained many queries that were not specific of any country. These queries are included in the dataset under the country label \texttt{none}.

\paragraph{Relevant documents} To find relevant passages for queries, we obtained Google Search ``featured snippets'' sourced from Wikipedia. These snippets are text excerpts displayed at the top of the search results in response to a query, deemed by Google to be relevant to the user's information need.\footnote{\href{https://blog.google/products/search/reintroduction-googles-featured-snippets/}{blog.google/products/search/reintroduction-googles-featured-snippets}}

To obtain snippets from Wikipedia, we appended the term ``wikipedia'' to each query before querying Google Search. We then extracted the entire Wikipedia paragraph associated with the featured snippet as the relevant document for each query. Queries not returning featured snippets from Spanish Wikipedia were discarded.

\paragraph{Corpus} We built a Spanish Wikipedia corpus with each paragraph corresponding to a single document. For this we used WikiExtractor \citep{attardi2015wikiextractor} to process the Spanish Wikipedia dump of 2024-04-01 and matched the extracted Wikipedia passages to the valid query-document pairs obtained from Google Search featured snippets.

The choice of Wikipedia as the dataset's collection was motivated by its accessibility, ease of processing, and the wide range of topics it covers.

\subsection{Data splits} \label{sec:data_splits}

\begin{table*}[ht]
  \centering
  \begin{tabular}{l c c c c c c c }
    \toprule
    \multicolumn{2}{c}{}        & \multicolumn{1}{c}{\textbf{Train}} & \multicolumn{5}{c}{\textbf{Test}}                                                         \\
    \cmidrule(lr){3-3} \cmidrule(lr){4-8}
    \multirow{-2.5}{*}{\textbf{Country}} & \multirow{-2.5}{*}{\textbf{Code}}   & \textbf{\thead{\#Unique \\ Q}}             & \textbf{\thead{\#Unique \\ Q}} & \textbf{\thead{\#Unique \\ Rel. Docs}} & \textbf{\thead{\#Unique \\ Rel. Articles}} & \textbf{\thead{Avg Q \\ Length}} & \textbf{\thead{Avg Rel. \\ Doc Length}} \\
    \midrule
    - & \texttt{full} & 537 730 & 156 528 & 63 557 & 47 127 & 5.7 & 79.9 \\
    \midrule
    - & \texttt{none} & 356 040 & 101 359 & 44 869 & 34 306 & 5.8 & 80.8 \\
    Argentina & \texttt{ar} & 22 560 & 5 481 & 3 829 & 3 498 & 5.4 & 80.4 \\
    Bolivia & \texttt{bo} & 24 912 & 4 810 & 3 230 & 2 866 & 5.3 & 79.7 \\
    Chile & \texttt{cl} & 22 486 & 5 408 & 3 694 & 3 381 & 5.4 & 79.3 \\
    Colombia & \texttt{co} & 25 914 & 5 667 & 3 845 & 3 464 & 5.6 & 79.8 \\
    Costa Rica & \texttt{cr} & 23 662 & 5 690 & 4 047 & 3 693 & 5.5 & 79.3 \\
    Cuba & \texttt{cu} & 22 071 & 4 787 & 3 374 & 3 071 & 5.4 & 80.9 \\
    Dominican Rep. & \texttt{do} & 27 830 & 5 359 & 3 725 & 3 320 & 5.6 & 79.8 \\
    Ecuador & \texttt{ec} & 27 599 & 6 074 & 4 214 & 3 734 & 5.9 & 81.1 \\
    Spain & \texttt{es} & 23 476 & 7 148 & 5 004 & 4 654 & 5.5 & 80.0 \\
    Guatemala & \texttt{gt} & 22 971 & 4 630 & 3 091 & 2 755 & 5.5 & 79.5 \\
    Honduras & \texttt{hn} & 26 818 & 5 608 & 3 890 & 3 540 & 5.5 & 78.8 \\
    Mexico & \texttt{mx} & 32 258 & 8 099 & 5 563 & 4 991 & 5.5 & 79.2 \\
    Nicaragua & \texttt{ni} & 28 179 & 5 787 & 3 926 & 3 539 & 5.4 & 78.8 \\
    Panama & \texttt{pa} & 25 784 & 5 777 & 4 082 & 3 700 & 5.5 & 79.8 \\
    Peru & \texttt{pe} & 25 877 & 5 458 & 3 784 & 3 443 & 5.5 & 79.6 \\
    Puerto Rico & \texttt{pr} & 26 609 & 6 343 & 4 347 & 3 930 & 5.4 & 81.2 \\
    Paraguay & \texttt{py} & 24 885 & 5 306 & 3 785 & 3 435 & 5.4 & 78.7 \\
    El Salvador & \texttt{sv} & 25 935 & 5 806 & 3 999 & 3 619 & 5.5 & 78.3 \\
    United States & \texttt{us} & 23 498 & 4 234 & 2 818 & 2 572 & 5.6 & 81.9 \\
    Uruguay & \texttt{uy} & 20 902 & 5 525 & 4 062 & 3 701 & 5.3 & 76.9 \\
    Venezuela & \texttt{ve} & 27 123 & 5 733 & 3 924 & 3 455 & 5.5 & 81.5 \\
    \bottomrule
\end{tabular}

  \caption{\textbf{The MessIRve dataset.}
    \#: number of, Q: queries, Rel.: relevant, Doc.: document. Queries come from Google's autocomplete API and documents are Wikipedia paragraphs. One paragraph may be relevant to multiple queries. The \texttt{full} set includes queries not specific to any country (code=\texttt{none}) plus the set of unique queries from all countries. Each query has one relevant document in each country. In the \texttt{full} set, a query may have multiple relevant documents. Statistics for the training set are nearly identical to those of the test set and are thus omitted.
  }
  \label{tab:dataset_splits}
\end{table*}

The dataset is split into training and test queries in such a way that the Wikipedia article to which any relevant paragraph belongs is present in only one of the splits. The partitioning was done by country, with about 20\% of the articles assigned to the test set. Splitting by article aims to reduce the overlap in topics between the training and test sets. We do not provide a validation or development set to allow users flexibility in their validation strategies, such as using cross-validation instead of relying on a single validation set.

Queries not specific to any country (label \texttt{none}) were combined with the set of unique queries from all countries to form a \texttt{full} set, which is also split into training and test sets. 
Unlike the country-specific sets, in the \texttt{full} set some queries can have multiple relevant documents because the same query may return different featured snippets in different country domains. 
Statistical information about the dataset splits is provided in Table \ref{tab:dataset_splits}.
Representative query-document examples for each country are provided in Appendix \ref{app:queries}, Table \ref{tab:dataset_examples}.

\subsection{Quality assessment} \label{sec:quality_assessment}

We evaluated the quality of the dataset using a method inspired by \citet{rybak-2023-maupqa}, sampling query-document pairs from the \texttt{full} test set and manually rating them on three binary criteria:

\begin{enumerate}[wide, itemindent=\labelsep, itemsep=0pt]
  \item \textbf{Query correctness:} Can the information need of the user be understood? This criterion can be considered met even if the query is not grammatically correct, which is the case for many queries issued to search engines. For example, the query ``porque se llama bogota'' (\emph{Why is it called Bogotá?}) can be considered correct even though it is missing question marks, accents, capitalization, etc.

  \item \textbf{Query unambiguity:} Can a good answer to the query be given without providing further detail? Raters were instructed to consider that the query may only make sense in the context of the user's country, which may not be explicitly stated. For example, the query ``cómo se llama el presidente'' (\emph{what is the name of the president}) is ambiguous without specifying the country, but it is clear that the user is asking about the president of their country, and so it can be considered unambiguous.

  \item \textbf{Document relevance:} Does the document help answer the query? Even if the passage does not provide a thorough answer, it can be deemed relevant if it contains information that, when combined with other sources, could help answer the query.
\end{enumerate}

Three independent native Spanish speakers hired via Prolific (\href{https://www.prolific.com/}{prolific.com}) rated the same 100 random samples for 13.60 USD/hr. Considering the majority vote, 93\% of the queries were rated as correct, 88\% as unambiguous, and 97\% of the documents were considered relevant to their queries. These results show that the dataset's quality is acceptable, as the queries are generally clear and the documents annotated as relevant are most likely to contain information that helps answer the query.

Because these three criteria are intrinsically ambiguous, i.e. they depend on each rater's interpretation, we measured the inter-rater agreement as the \% of the queries on which all annotators agree. We obtained 80\% for correctness, 65\% for unambiguity, and 83\% for relevance. These results support the reliability of the quality assessment.

To further assess the quality of the dataset, we used GPT-4o and Claude-3.5-Sonnet to annotate the relevance of other 650 query-document pairs. They identified 86\% and 93.4\% of the documents as relevant, respectively, with an agreement in 91.1\% of the samples. To better understand the results, we manually reviewed the cases where both models marked documents as non-relevant. No clear pattern emerged from this analysis. Refer to Appendix \ref{app:quality_assessment} for examples.

Finally, as an additional validation, four of the paper's authors carried out a quality assessment of these 650 pairs.
We obtained 93.5\% for correctness, 90.1\% for unambiguity, and 90.2\% for relevance.
The consistence across independent annotators, coauthors, and LLM judges highlights the robustness of the dataset's quality.
Further details and a discussion of inter-rater agreement can be found in Appendix~\ref{app:quality_assessment}.

\section{Comparison with existing datasets} \label{sec:comparison}

MessIRve provides detailed documentation of its data collection process, unlike existing open-domain datasets like MIRACL and SQAC, which lack specifics on Wikipedia passage selection. Our dataset is also the only one to account for Spanish dialectal diversity, whereas other datasets either ignore this or fail to document it. To validate this, we identified distinctive words in each country's queries using log odds ratios (Table \ref{tab:spanish_varieties} in Appendix \ref{app:queries}). Examples for the countries with most speakers include ``green'' (USA, Green Card), ``checar'' (Mexico, ``to check''), ``frailejones'' (Colombia, plant), ``euskera'' (Spain, Basque language), and ``CUIT'' (Argentina, tax ID).

MessIRve is also much larger than other datasets (Table \ref{tab:current_datasets}). While mMARCO is large, it relies on translations of English-speaking users' queries, introducing translation errors and contents less relevant to Spanish-speaking users (see Section \ref{sec:topic_analysis}).

Like MSMARCO, MessIRve has sparse judgments, with only about one per query. It is thus more likely to have more false negatives compared to denser datasets like MIRACL. Additionally, MessIRve lacks explicitly negative judgments, requiring negative mining techniques for training (Section \ref{sec:experiments}). IR systems trained using our dataset would benefit from more complete annotations.

Despite this, sparse and positive-only judgments are common in IR due to the challenges and high costs of annotation \citep{lin2022pretrained}. 
Explicit negative annotations are rare; for example, only one dataset in BEIR has them (Touché-2020) and they are not included in the benchmark \citep{thakur2021beir}. In addition, the number of negative annotations per query is usually small, e.g. MIRACL has 5.3 negative annotations per query, so some negative sampling is still required for training.
Previous research shows that sparse judgments, like MSMARCO's, yield rankings comparable to those from more comprehensive and costly annotations \citep{zhang2021mr}.

MessIRve remains valuable, especially within a broader evaluation framework combining datasets with varying judgment densities, as seen in BEIR \citep{thakur2021beir}. Our fine-tuning experiments (Section \ref{sec:experiments}) also show that BM25-mined hard negatives improve performance; more advanced sampling strategies could further improve results.

Our automated data collection prioritized scale and avoiding costly manual annotations, like in MIRACL and MSMARCO. This sacrifices some control over quality, relying on automated query extraction and the accuracy of Google Search snippets. For this reason, we assessed the dataset's reliability in Section \ref{sec:quality_assessment}. 
In the following sections we compare the datasets in terms of their format and topics.


\subsection{Query format} \label{sec:queries_format}

\begin{figure}[ht]
  \centering
  \includegraphics[width=0.9\columnwidth]{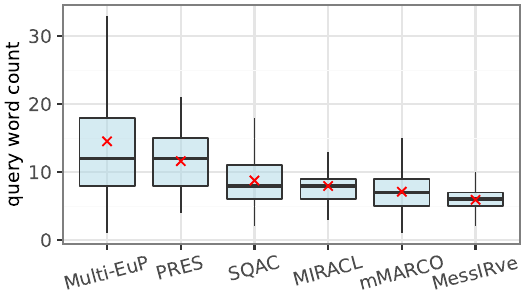}
  \caption{\textbf{Distribution of query lengths in each dataset.}
  Boxplots exclude outliers for better visualization. Red crosses indicate the mean of each distribution.
  }
  \label{fig:datasets_query_len}
\end{figure}

MessIRve queries are generally shorter and less variable in length than those in other datasets (Figure \ref{fig:datasets_query_len}).
The shorter length of MessIRve queries is likely a characteristic of search engine queries, in contrast to datasets such as MIRACL or SQAC, which are collected from human annotators.

Short queries are not necessarily simple or generic. To show this, we computed the \% of queries starting with typical question words in open-domain datasets (Spanish for ``who'', ``what'', ``which'', ``when'', ``where'', ``how many'', ``how much'', ``why'', ``how'').
We obtained 65.5\% for MessIRve, 92\% for MIRACL, 82.2\% for SQAC, and 59.7\% for mMARCO.
This suggests that MessIRve is not disproportionately biased toward simple or generic questions when compared to other open-domain IR datasets.
Examples of MessIRve queries not starting with these question words include ``para qué sirve'' (\emph{what is ... for}), ``para qué es'' (\emph{what is ... for}), and ``en qué consiste ...'' (\emph{what does ... consist of}).
These account for 2.33\% of MessIRve queries but less than 0.30\% in other datasets.

To further examine query structure, we measured prefix diversity using the entropy for prefix lengths from 1 to 6 (Appendix \ref{app:queries}, Figure \ref{fig:datasets_query_entropy}).
MessIRve and mMARCO show the highest entropy for multi-word prefixes, indicating greater variability in question openings.
For single-word prefixes, MEUP ranks highest, which is consistent with its queries being debate topics.


\subsection{Analysis of topics} \label{sec:topic_analysis}


MessIRve's scale allows for a wider variety of topics than other datasets. For example, the \texttt{full} set of MessIRve contains judged passages coming from 84,284 unique Wikipedia articles, while MIRACL covers 17,640 articles and SQAC, 3,823 articles (considering training, validation and test sets).

\begin{figure*}[ht]
  \includegraphics[scale=0.75]{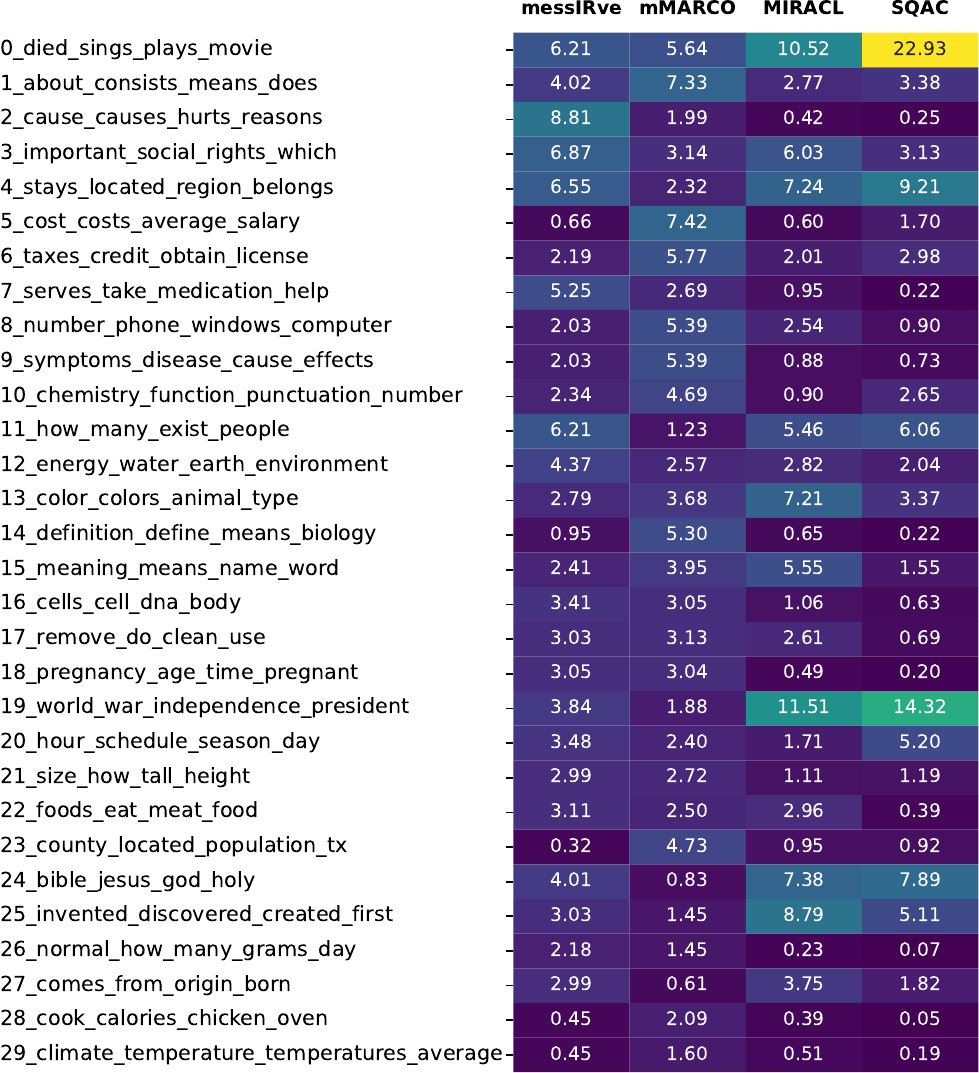}
  \centering
  \caption{\textbf{Topics in Spanish IR datasets.} Topics are sorted by overall frequency from top to bottom, and are labeled with an ID and the four most representative words according to their c-TF-IDF values (see \citealp{grootendorst2022bertopic}). Numbers in the heatmap represent the \% of queries in a dataset that belong to a topic. The color scale is shared across datasets. For the sake of clarity, words were translated from Spanish to English by the authors.
  }
  \label{fig:datasets_topics}
\end{figure*}

To more thoroughly analyze the topics covered by the queries of the different datasets, we employed BERTopic with PCA of 30 dimensions of \texttt{multilingual-e5-large} embeddings, and KMeans clustering with 30 clusters \citep{grootendorst2022bertopic,wang2024multilingual}. We assume just one topic per query, and focus only on the queries of the general-domain datasets: MIRACL, SQAC, mMARCO, and MessIRve \texttt{full}. For every dataset we use the complete set of available queries.

The clusters were labeled with the four most representative words according to their c-TF-IDF values \citep{grootendorst2022bertopic}. To provide a summary of a particular topic in a given dataset, representative queries were found by sampling a subset of queries from the dataset belonging to the topic and calculating which ones were closest to the topic's c-TF-IDF representations based on cosine similarity. Representative words of a topic in a dataset were also found by considering the highest c-TF-IDF scores of the words from the dataset.

The distribution of topics accross datasets (Figure \ref{fig:datasets_topics}) reveals two broad miscellaneous topics which are relatively prevalent in all datasets: queries concerning entertainment celebrities, including those in sports, music, movies, and other media (topic 0) and general definitions (topic 1).

MIRACL and SQAC have more topics on war and politics (topic 19), religion (topic 24), and inventions or discoveries (topic 25), likely due to their Wikipedia-based construction. Relative to other datasets, MessIRve focuses on health-related pain (topic 2), while mMARCO features household economy (topics 5 and 6) and technology (topic 8).

It is interesting that topic 23, prevalent in mMARCO but not MessIRve, concerns US demographics, with representative queries like ``dallas nc se encuentra en qué condado'' (\emph{Dallas, NC is located in what county}) or ``en qué condado se encuentra grove city ohio'' (\emph{In what county is Grove City, Ohio located}). In contrast, topic 4, relevant in MessIRve, SQAC, and MIRACL, focuses on broader demographics, with queries such as ``donde queda logroño'' (\emph{where is Logroño}), or ``por donde queda montañita'' (\emph{where is Montañita}), reflecting Spanish-speaking users' needs more than topic 23 of mMARCO, whose queries probably come from English-speaking users.

Similarly, topic 0 in MessIRve and mMARCO highlights popularity differences in Spanish-speaking regions. mMARCO queries include ``en qué equipo juega michael sam'' (\emph{what team does Michael Sam play for}), referring to a US NFL player, whereas MessIRve has ``en cuál equipo juega james rodriguez'' (\emph{which team does James Rodriguez play for}), referencing a Colombian soccer player. This aligns with MessIRve's extraction of queries by Spanish-speaking country, unlike mMARCO's translated English queries.

MessIRve and mMARCO have a more balanced topic distribution than MIRACL and SQAC, where certain topics dominate in frequency. This likely results from their larger size, allowing for greater topic diversity. However, MIRACL and SQAC contain well-formed questions, whereas MessIRve has more colloquial queries, better representing real user searches. mMARCO also shares this trait but is influenced by translation quality. Table \ref{tab:datasets_topics} in Appendix \ref{app:queries} provides representative words and queries for the most frequent topics.

We stress that MessIRve's topic analysis is not a faithful representation of Spanish-speaking users' real interests, but rather a descriptive overview of the dataset composition compared to others.

\section{Experiments} \label{sec:experiments}

As a baseline, we evaluate IR models on MessIRve's and other IR datasets' test sets. 
We consider the following models for evaluation:

\begin{itemize}[wide, itemindent=\labelsep, itemsep=0pt]

  \item \textbf{BM25} \citep{robertson1994okapi}, a popular lexical IR model based on the token overlap between query and documents. Despite its simplicity, BM25 has proven to be a strong baseline \citep{thakur2021beir}.

  \item \textbf{MIRACL-mdpr-es} \citep{zhang2023miracl}. This is a bi-encoder initalized from multilingual Dense Passage Retriever (mDPR, \citealp{zhang-etal-2021-mr}), which was first pre-fine-tuned using the training set of MSMARCO and further fine-tuned on the MIRACL Spanish training set. We use the \texttt{mdpr-tied-pft-msmarco-ft-miracl-es} checkpoint.\footnote{\href{https://huggingface.co/castorini/mdpr-tied-pft-msmarco-ft-miracl-es}{huggingface.co/castorini/mdpr-tied-pft-msmarco-ft-miracl-es}} As far as we know, this is the only well-documented, Spanish-specific retriever available.

  \item \textbf{E5-large} \citep{wang2024multilingual}. This bi-encoder is initialized from the multilingual model XLM-R \citep{conneau-etal-2020-unsupervised} and then trained with weak supervision in a mixture of multilingual text pairs, followed by supervised fine-tuning in annotated data which is mostly in English but includes some multilingual data e.g. MIRACL. We use the \texttt{multilingual-e5-large} checkpoint.\footnote{\href{https://huggingface.co/intfloat/multilingual-e5-large}{huggingface.co/intfloat/multilingual-e5-large}}

  \item \textbf{OpenAI-large}. Although the architecture and training data are not publicly disclosed, previous OpenAI models \citep{neelakantan2022text} consist of a bi-encoder that maps documents and query to embeddings, and is trained with contrastive learning on several supervised datasets. We use \texttt{text-embedding-3-large}  via an API.\footnote{\href{https://platform.openai.com/docs/guides/embeddings}{platform.openai.com/docs/guides/embeddings}}

  \item \textbf{E5-large-ft-messirve}. We fine-tuned an E5-large model on the MessIRve \texttt{full} training set, retrieving hard negatives with BM25 and following the same approach as \citet{wang2024multilingual}. Refer to Appendix \ref{app:fine_tuning} for more details on the training.

\end{itemize}

In all cases, we append the title of the Wikipedia article to the document text before retrieval. Except for E5-large-ft-messirve, all evaluations are conducted in a zero-shot manner.

We use two standard metrics for evaluation \citep{thakur2021beir, zhang2023miracl}:

\begin{itemize}[wide, itemindent=\labelsep, itemsep=0pt]
  \item \textbf{Recall@100:} the fraction of relevant documents within the top 100 results, averaged over all queries.
  \item \textbf{nDCG@10:} the normalized Discounted Cumulative Gain. It compares the rank of the top 10 results to the ideal ranking  where relevant documents are ranked higher. It is averaged over all queries.
\end{itemize}

\begin{figure}[ht]
  \centering
  \includegraphics[width=\columnwidth]{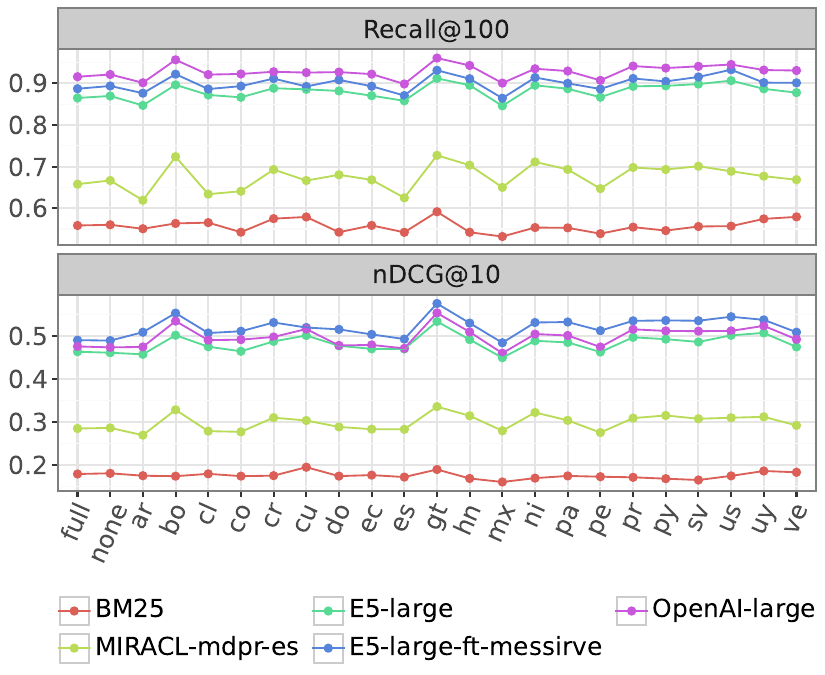}
  \caption{\textbf{Baseline results on MessIRve test sets.} Metrics include nDCG@10 and Recall@100. All dense models consistently outperform BM25. Fine-tuned E5-large achieves the highest nDCG@10 across all subsets, while zero-shot OpenAI-large leads in Recall@100.
  }
  \label{fig:models_results}
\end{figure}

\begin{figure}[ht]
  \centering
  \includegraphics[width=\columnwidth]{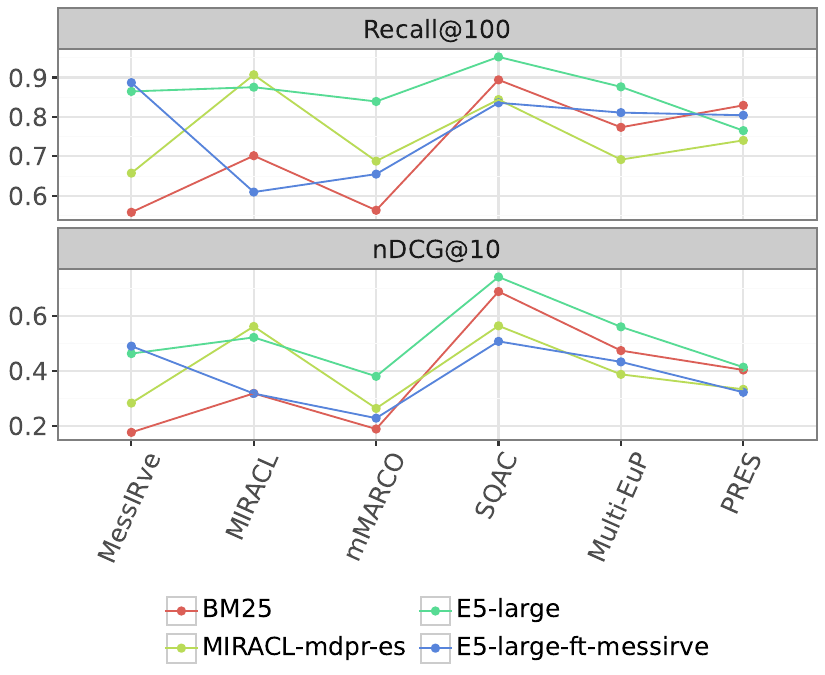}
  \caption{\textbf{Baseline results on Spanish IR datasets.} We consider only open-source models due to the cost of evaluating OpenAI-large. We use the test queries of each dataset except MIRACL, for which we use the dev set, since the test set is not publicly available.
  }
  \label{fig:other_datasets_results}
\end{figure}

Figure \ref{fig:models_results} shows the results of the models on the MessIRve test split in terms of nDCG@10 and Recall@100, segmented by country plus the \texttt{full} and \texttt{none} sets. The OpenAI-large and E5-large models outperform smaller models (MIRACL-mdpr-es) and the lexical model (BM25) across all dataset partitions. In terms of nDCG@10, OpenAI-large achieves an average score in the \texttt{full} set of $.476$ compared to E5-large's $.463$, indicating a slightly higher zero-shot ranking effectiveness. For Recall@100, OpenAI-large consistently outperforms E5-large accross the dataset partitions, with a recall of $.916$ versus $.865$ in the \texttt{full} set.

The fine-tuned version of E5-large on MessIRve surpasses all other zero-shot models in terms of nDCG@10 across all subsets, achieving a score of $.491$ in the \texttt{full} set. This means that, even if the queries in the test set refer to articles that are not in the training set, the fine-tuned model is able to learn a better ranking function. While the fine-tuning improves the zero-shot recall of E5-large to $.887$, it does not surpass the recall of OpenAI-large.

BM25 consistently performs the worst across all subsets (nDCG@10=$.179$ and Recall@100=$.558$ in the \texttt{full} set), highlighting the limitations of lexical methods compared to dense models, particularly in queries written in colloquial language.

Although MIRACL-mdpr-es is specifically fine-tuned for Spanish, it lags behind the larger models, with nDCG@10 and Recall@100 scores of .284 and .658, respectively, suggesting that model size and fine-tuning strategy might play a more critical role than language-specific tuning. For example, while MIRACL-mdpr-es has 178M parameters and is pre-fine-tuned with MSMARCO and then fine-tuned with Spanish MIRACL, E5-large has 560M parameters, is pre-trained in a mixture of multilingual text pairs, and then fine-tuned in at least 10 supervised datasets, including MSMARCO and multilingual data like MIRACL itself.

All dense models follow a similar performance pattern, with the highest scores in the \texttt{bo} (Bolivia) and \texttt{gt} (Guatemala) subsets and the lowest in \texttt{ar} (Argentina) and \texttt{mx} (Mexico).

We also compare the methods across the datasets described in Section \ref{sec:related_work} (Figure \ref{fig:other_datasets_results}). 
The model rankings vary depending on the dataset, based on both Recall and NDCG. Overall, E5-large performs relatively well across all datasets. 
In SQAC, Multi-EuP, and PRES, BM25 also shows strong results, suggesting that lexical overlap is important in these datasets, unlike the others. 
The fine-tuned models, MIRACL-mdpr-es and E5-large-ft-messirve, perform best only in their respective datasets but struggle in others. 
This suggests that each dataset is quite different from the others.
It also indicates that individual training datasets alone may not be enough for achieving generalization to other domains through fine-tuning, but they could be useful when combined with other training datasets.

Finally, to better understand the dataset's limitations, we conducted a manual error analysis on 30 queries where the top-performing retriever, OpenAI-large, failed to find a relevant passage within the top-20 results.
For each query, we reviewed the annotated relevant passages alongside the top 5 retrieved passages.
We found that 56.7\% (17/30) of the cases included false negatives due to non-exhaustive annotations, 23.3\% (7/30) came from ambiguous queries, and 13.3\% (4/30) had documents erroneously marked as relevant.
This means that errors can come from dataset issues, as well as from models' capabilities.

\section{Discussion and conclusions} \label{sec:conclusions}

We presented MessIRve, a large-scale dataset for information retrieval in Spanish. MessIRve queries originate from Spanish speakers from almost all Spanish-speaking countries, ensuring relevance to the information needs of the Spanish-speaking community. This contrasts with other datasets, which are either translated from English or do not consider dialectal variations. Moreover, the dataset's large scale allows it to cover a wide variety of topics, unlike smaller datasets.

We performed a quality assessment to ensure the reliability of the dataset, and we provided a detailed comparison with existing Spanish IR datasets, highlighting the dataset's strengths and limitations. We also provide baseline results of the performance of various IR models on the datasets.

It should be noted that the dataset was collected using an IR system, Google's ``featured snippets''. Although there's no certainty about how it works, it's likely that it considers many other factors besides the text of the query and the documents, such as inbound links, user behavior, the structure of the web, etc., implying significant computational resources. This probably contributes to the high quality of the dataset, making it a valuable resource for training and evaluating open-source IR models. 

However, relying on Google's system introduces potential biases.
Our dataset may reflect Google's ranking priorities and content preferences, and is unsuitable for identifying IR systems that outperform Google's own technology.
Like all IR datasets, ours has limitations in terms of scope, collection methods and underlying sources.
To mitigate such biases, we advocate evaluating IR models using a diverse set of datasets rather than just one, as demonstrated by benchmarks such as BEIR \citep{thakur2021beir}.
By contributing to dataset diversity and complementing existing resources, MessIRve supports the broader goal of building a comprehensive Spanish-language IR benchmark. 

We have focused on zero-shot baselines and fine-tuning on the full set. Future work includes fine-tuning on country-specific subsets, and creating a unified Spanish IR benchmark with all datasets.

\section*{Limitations}

The dataset collected may not be representative of the distribution of popular topics in each country, as it only includes a sample of popular queries in Spanish-speaking countries at the time of collection. Moreover, it only consists of queries which have an answer in Spanish Wikipedia, and that Google is confident in answering.

While we made an effort to ensure that the dataset is free from offensive content, we did not conduct a systematic analysis. Instead, we checked for the absence of popular offensive terms in Spanish, which may have overlooked some inappropriate content.

Despite our efforts to include different Spanish-speaking regions, certain dialects or linguistic variations may still be underrepresented in the queries of the dataset. Reliance on Wikipedia as a source of documents may also introduce bias, as Wikipedia content is far from comprehensive in terms of topics and viewpoints, and may exclude topics of importance to some communities.

To examine this issue further, we analyzed Wikipedia edit statistics, specifically the average distribution of monthly active editors from 2018 to 2024. The countries contributing more than 1\% include \texttt{es} (28.1\%), \texttt{ar} (13.1\%), \texttt{mx} (12.4\%), \texttt{cl} (8.0\%), \texttt{co} (7.4\%), \texttt{pe} (6.3\%), \texttt{us} (2.4\%), \texttt{uy} (2.1\%), \texttt{ec} (2.1\%), \texttt{cr} (1.2\%), \texttt{bo} (1.1\%).\footnote{\href{https://dumps.wikimedia.org/other/geoeditors/readme.html}{dumps.wikimedia.org/other/geoeditors/readme.html}} This distribution suggests that while contributions are skewed towards Spain, it still reflects representation from various Spanish-speaking countries, but still excluding some.

Although our dataset represents an improvement over what is currently available in Spanish, IR systems trained on this dataset could perpetuate existing biases from the queries, corpus selection, and relevance annotations of Google's ``featured snippets''. For example, retrieval results could prioritize topics or perspectives from well-represented communities, while underrepresenting minority viewpoints.

We did not assess the statistical significance of our results or compute confidence intervals. However, given the dataset's scale, we believe that the performance estimates provided are reliable.

Finally, the dataset released should not be used for commercial purposes.

\section*{Acknowledgments}

This work used computational resources from CCAD -- Universidad Nacional de Córdoba (\url{https://ccad.unc.edu.ar/}), which are part of SNCAD -- MinCyT, República Argentina, and from NodoIA San Francisco (Ministry of Science and Technology of the Province of Córdoba, Argentina).

We thank: OpenAI for their financial support to run experiments with their models; authors of GooAQ \citep{khashabi-etal-2021-gooaq-open} and Pyserini \citep{Lin_etal_SIGIR2021_Pyserini} for publicly sharing their code, which significantly contributed to our research; and Chris Petrillo of Wikipedia for generously providing Wikipedia edit statistics.

\bibliography{anthology,custom}

\appendix

\section{Inference and fine-tuning} \label{app:fine_tuning}

To evaluate the performance of BM25, we indexed the corpus and ran retrieval using Pyserini with default parameters and Spanish analyzer \citep{Lin_etal_SIGIR2021_Pyserini}.

Inference and training of dense retrieval models was performed with the \texttt{transformers} library version 4.35.2 \citep{wolf2020transformers} on 2 NVIDIA A30 GPUs with 24GB of memory each. We used the \texttt{deepspeed} library \citep{rasley2020deepspeed}, which allows for efficient training of large models.

To fine-tune the E5-large model on MessIRve, we followed \citet{wang2024multilingual} and \citet{wang2024text}, and used the following hyperparameters: we train for 2 epochs using a batch size of 128, a learning rate of $1e^{-5}$ with linear decay and the first 400 steps for warmup, and we used 7 hard negatives extracted with BM25 for each positive example.

To reduce GPU memory usage, we used mixed precision training, gradient checkpointing, and truncated queries and documents to a maximum length of 64 and 256 tokens, respectively. We also keep position embeddings frozen during fine-tuning.

It took around 15 hours to fine-tune the model on the MessIRve \texttt{full} training set. No hyperparameter search was performed.

\section{Quality assessment} \label{app:quality_assessment}

To provide additional support for the quality assessment of the dataset, four of the authors of the paper, native Spanish speakers from Argentina, evaluated a larger number of samples, trying to be as objective as possible and working independently of each other. Each of them rated 200 query-document pairs, with 50 pairs in common across all raters, for a total of 650 unique pairs.

93.5\% of the queries were rated as correct, 90.1\% were rated as unambiguous, and 90.2\% of the documents were considered relevant to their queries, using the majority vote for overlapping queries. These results validate the results obtained with the hired independent raters. We also measured the inter-rater agreement by computing the \% of the 50 common queries on which all annotators agree.  We obtained 78\% for query correctness, 74\% for query unambiguity, and 78\% for document relevance. These results further support the reliability of the quality assessment.

To better understand inter-rater agreement, we show the distribution of positive ratings across queries for each aspect, based on assessments by Prolific raters:

\begin{itemize}[wide, nosep, itemindent=\labelsep]
    \item Correct.: {3 raters: 80\%, 2: 13~\%, 1: 7\%, 0: 0\%}
    \item Non-ambiguity: {3: 64\%, 2: 24\%, 1: 11\%, 0: 1\%}
    \item Relevance: {3: 83\%, 2: 14\%, 1: 3\%, 0: 0\%}
\end{itemize}

This shows that full agreement on non-relevance is rare: all raters never simultaneously assign a negative rating. This suggests that when a passage is rated as non-relevant by some rater, it might be due to ambiguity rather than clear irrelevance.

Additionally, we see that because samples appear to be generally good (positives are the majority class), there are few samples left to evaluate what the agreement might be in the negative class. 
This imbalance can inflate simple \% agreement: when the marginal probability of assigning a positive label is high, \% agreement can be high even if ratings are independent.

To address this, we computed Fleiss' kappa ($\kappa$) in the Prolific-rated samples: 
we obtained 0.186 for correctness, 0.146 for non-ambiguity, and 0.089 for relevance, indicating slight agreement.
However, given the high marginal probability and the sample size, the variance of $\kappa$ is high, requiring a larger sample for reliable measurement. 
Intuitively, when most samples belong to one class, fewer are available to assess actual agreement. 

In our analysis of instances annotated as non-relevant by GPT-4o and Claude-3.5-Sonnet, we found some queries that were inherently ambiguous, making it difficult to determine whether the retrieved document was relevant. Examples include: 

\begin{itemize}[wide, itemindent=\labelsep, itemsep=0pt]
    \item ``de cuántos grados fue el temblor de hoy en Colombia'' (\emph{how many degrees was today's earthquake in Colombia}). The document retrieved discusses the Urrao Earthquake of 2021 which we know is not relevant because of the time frame of the query.
    
    \item ``quién se queda con el perro'' (\emph{who gets the dog}). The passage retrieved is about ``¿Con quién se queda el perro?'', the title of a music album, rather than addressing a legal or situational question regarding pet ownership.
    
    \item ``qué hace crema para peinar'' (\emph{what hair styling cream do}). This phrase is grammatically incorrect and lacks specificity, making it difficult to determine the correct interpretation.

    \item ``por qué las articulaciones suenan'' (\emph{why do joints/articulations make a sound}). The passage retrieved discusses phonetics, while the query is more likely referring to the physiological phenomenon of joint cracking.
\end{itemize}

Other queries seem to be too general to be accurately matched with a specific document, for example ``cuáles son derechos económicos'' (\emph{which are economic rights}); the query is broad and may refer to various aspects of economic rights, making it challenging to determine whether a given document sufficiently addresses the intended scope.

We observed cases of clear mismatch where the retrieved passage was clearly non-relevant to the query. Examples include:

\begin{itemize}[wide, itemindent=\labelsep, itemsep=0pt]
    \item ``por qué es rosa la sal del Himalaya'' (\emph{why is Himalayan salt pink}). The passage retrieved discusses black salt and explicitly states that it should not be confused with pink Himalayan salt.
    
    \item ``en qué isla se filmó La Laguna Azul'' (\emph{on which island was The Blue Lagoon filmed}). The passage retrieved describes Blue Lagoon Island in the Bahamas but does not explicitly confirm whether this was the filming location of ``La Laguna Azul''.
    
\end{itemize}

Finally, we identified instances where the negative annotation assigned by the LLMs might be debatable. For example, for the query ``qué hizo John Wesley'' (\emph{what did John Wesley do}), the passage states that John Wesley was famous for organizing the 1869 expedition through the Green and Colorado Rivers in Utah. While this passage does not provide a comprehensive biography, it still presents a relevant historical action associated with John Wesley, making its exclusion questionable.

These findings highlight potential areas for refinement in future dataset evaluations.

\section{Queries} \label{app:queries}

\begin{figure}[ht!]
  \centering
  \includegraphics[height=11cm]{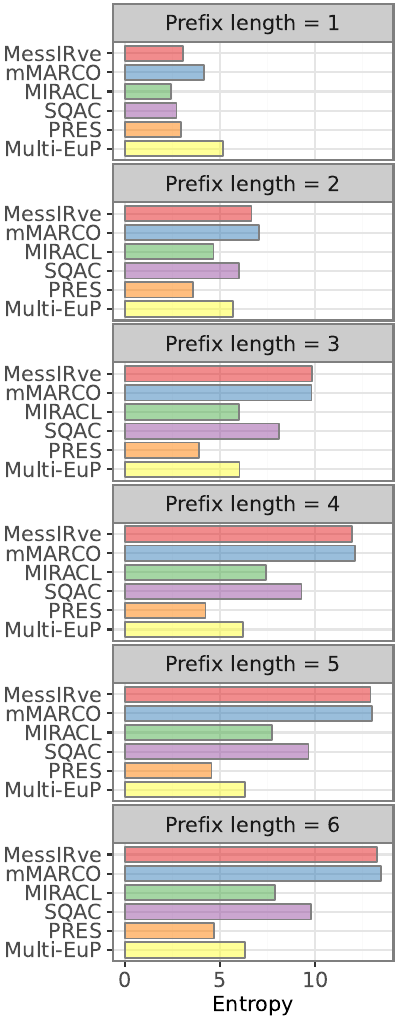} 
  \includegraphics[height=11cm]{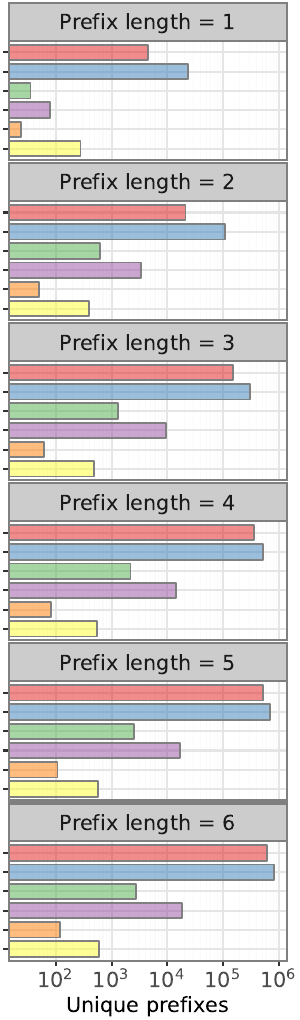}
  \caption{\textbf{Entropy and number of unique prefixes of length 1 to 6 in each dataset.}
    The entropy measures the diversity of the prefix distribution, with higher values indicating more variability. The number of unique prefixes indicates how many different prefixes of a given length exist in the dataset.
  }
  \label{fig:datasets_query_entropy}
\end{figure}

Table \ref{tab:google_prefixes} shows the seed prefixes used to obtain the queries for each country in the MessIRve dataset. These are loosely based on \citet{khashabi-etal-2021-gooaq-open}. We include incorrect, non-accented, versions of words, e.g. ``como'' instead of ``cómo'', to account for common spelling mistakes in Spanish queries.

The 120 prefixes used are \emph{seed} prefixes. For each country, we send each seed prefix to the API with all one-letter continuations (1+27 API calls per prefix). We compile results, retain those starting with a seed prefix, and generate new prefixes adding 1 word. Then we consider 2-word continuations, and stop when reaching a target results count per country. The number of API results per call varies: more specific prefixes often yield fewer results. We do not extract a fixed number of completions per prefix, and we retain queries not starting with seed prefixes (but don't use them for further API calls). 

Table \ref{tab:spanish_varieties} shows the most distinctive words for each country in the MessIRve queries. These words are obtained by calculating the log odds ratio of the frequency of each word in the queries for each country with respect to the queries of the \texttt{full} subset, and selecting the words with the highest scores. The words are indicative of the dialectal variety of Spanish spoken in each country.

Table \ref{tab:datasets_topics} shows the most representative words and queries for the 8 most frequent topics in each dataset. Representative words and queries were selected based on the c-TF-IDF scores \citep{grootendorst2022bertopic}.

Representative query-document pairs per country are shown in Table \ref{tab:dataset_examples}.
Queries are sampled from those containing the top log-odds ratio words per country, as listed in Table \ref{tab:spanish_varieties}.

MessIRve contains shorter queries than other datasets, but with higher prefix diversity. This is shown in Figure \ref{fig:datasets_query_len} (distribution of query lengths in each dataset) and Figure \ref{fig:datasets_query_entropy} (entropy and number of unique prefixes of length 1 to 6 in each dataset).


\begin{table*}[ht!]
  \small
      \texttt{``a cual'', ``a cuales'', ``a cuantas'', ``a cuantos'', ``a cuál'', ``a cuáles'', ``a cuántas'', ``a cuántos'', ``a donde'', ``a dónde'', ``a que'', ``a quien'', ``a quienes'', ``a quién'', ``a quiénes'', ``a qué'', ``adonde debe'', ``adónde debe'', ``ante cual'', ``ante cuales'', ``ante cuál'', ``ante cuáles'', ``ante que'', ``ante quien'', ``ante quienes'', ``ante quién'', ``ante quiénes'', ``ante qué'', ``bajo cual'', ``bajo cuales'', ``bajo cuál'', ``bajo cuáles'', ``bajo que'', ``bajo qué'', ``buenas razones para'', ``buenos motivos para'', ``como'', ``con cual'', ``con cuales'', ``con cuanta'', ``con cuantas'', ``con cuanto'', ``con cuantos'', ``con cuál'', ``con cuáles'', ``con cuánta'', ``con cuántas'', ``con cuánto'', ``con cuántos'', ``con que'', ``con quien'', ``con quienes'', ``con quién'', ``con quiénes'', ``con qué'', ``cual'', ``cuales'', ``cuando'', ``cuanta'', ``cuantas'', ``cuanto'', ``cuantos'', ``cuál'', ``cuáles'', ``cuándo'', ``cuánta'', ``cuántas'', ``cuánto'', ``cuántos'', ``cómo'', ``de cual'', ``de cuanta'', ``de cuantas'', ``de cuanto'', ``de cuantos'', ``de cuál'', ``de cuánta'', ``de cuántas'', ``de cuánto'', ``de cuántos'', ``de donde'', ``de dónde'', ``de que'', ``de quien'', ``de quienes'', ``de quién'', ``de quiénes'', ``de qué'', ``deberia'', ``deberian'', ``desde donde'', ``desde dónde'', ``desde que'', ``desde qué'', ``donde'', ``durante cual'', ``durante cuales'', ``durante cuanto'', ``durante cuantos'', ``durante cuál'', ``durante cuáles'', ``durante cuánto'', ``durante cuántos'', ``durante que'', ``durante qué'', ``dónde'', ``en cual'', ``en cuales'', ``en cuantas'', ``en cuanto'', ``en cuantos'', ``en cuál'', ``en cuáles'', ``en cuántas'', ``en cuánto'', ``en cuántos'', ``en donde'', ``en dónde'', ``en que'', ``en quien'', ``en quién'', ``en qué'', ``entre cuales'', ``entre cuáles'', ``entre que'', ``entre quienes'', ``entre quiénes'', ``entre qué'', ``hacia cual'', ``hacia cuál'', ``hacia que'', ``hacia quien'', ``hacia quién'', ``hacia qué'', ``hasta cual'', ``hasta cuando'', ``hasta cuanta'', ``hasta cuantas'', ``hasta cuanto'', ``hasta cuantos'', ``hasta cuál'', ``hasta cuándo'', ``hasta cuánta'', ``hasta cuántas'', ``hasta cuánto'', ``hasta cuántos'', ``hasta donde'', ``hasta dónde'', ``hasta que'', ``hasta qué'', ``mediante cual'', ``mediante cuales'', ``mediante cuál'', ``mediante cuáles'', ``mediante que'', ``mediante qué'', ``motivos de por que'', ``motivos de por qué'', ``motivos de'', ``motivos del'', ``motivos para'', ``motivos por los que'', ``motivos por los qué'', ``para cual'', ``para cuantas'', ``para cuantos'', ``para cuál'', ``para cuántas'', ``para cuántos'', ``para donde'', ``para dónde'', ``para que'', ``para quien'', ``para quienes'', ``para quién'', ``para quiénes'', ``para qué'', ``podria'', ``por cuanta'', ``por cuantas'', ``por cuanto'', ``por cuantos'', ``por cuánta'', ``por cuántas'', ``por cuánto'', ``por cuántos'', ``por donde'', ``por dónde'', ``por que debe'', ``por que deberia'', ``por que no'', ``por que'', ``por qué debe'', ``por qué deberia'', ``por qué no'', ``por qué'', ``porque'', ``porqué'', ``pros y contras de'', ``puede'', ``pueden'', ``que'', ``quien'', ``quienes'', ``quién'', ``quiénes'', ``qué'', ``razones de por que'', ``razones de por qué'', ``razones de'', ``razones del'', ``razones para'', ``razones por las que'', ``razones por las qué'', ``segun quien'', ``segun quienes'', ``segun quién'', ``segun quiénes'', ``según quien'', ``según quienes'', ``sobre cual'', ``sobre cuál'', ``sobre que'', ``sobre quien'', ``sobre quienes'', ``sobre quién'', ``sobre quiénes'', ``sobre qué'', ``tras cuantas'', ``tras cuanto'', ``tras cuantos'', ``tras cuántas'', ``tras cuánto'', ``tras cuántos''} \\

  \caption{\textbf{Seed prefixes used to obtain queries from the Google autocomplete API.}
  }
  \label{tab:google_prefixes}
\end{table*}

\begin{table*}[ht!]
  \centering
  \begin{tabular}{lcccccc}
    \toprule
    \texttt{ar} & antula & 1982 & cuit & pampita & renga & rivadavia \\ 
    \midrule
    \texttt{bo} & samaipata & bulo & pagador & willka & katari & evo \\ 
    \midrule
    \texttt{cl} & huilo & pichilemu & pasamos & apuros & chiloe & colocolo \\ 
    \midrule
    \texttt{co} & eliecer & gaitan & covenas & nevados & frailejones & caldas \\ 
    \midrule
    \texttt{cr} & puntarenas & alajuela & bifonazol & guanacaste & rica & hidroxizina \\ 
    \midrule
    \texttt{cu} & maceo & cienfuegos & baragua & monilia & yate & bayamo \\ 
    \midrule
    \texttt{do} & enriquillo & bani & balaguer & luperon & dominicano & altagracia \\ 
    \midrule
    \texttt{ec} & montubios & veintimilla & guayasamin & proponen & ecuatoriana & roldos \\ 
    \midrule
    \texttt{es} & croquetas & declinaciones & protestan & supervivientes & naiara & euskera \\ 
    \midrule
    \texttt{gt} & mazatenango & monterrico & pacaya & tikal & verapaz & xincas \\ 
    \midrule
    \texttt{hn} & sula & tegucigalpa & morazan & copan & lempira & yojoa \\ 
    \midrule
    \texttt{mx} & maluma & decanato & redactaron & kilataje & kumbia & checar \\ 
    \midrule
    \texttt{ni} & masaya & gueguense & matagalpa & sandino & managua & esteli \\ 
    \midrule
    \texttt{pa} & delfia & panamena & cortez & rolo & martinelli & torrijos \\ 
    \midrule
    \texttt{pe} & boluarte & quispe & chiclayo & benavides & chimbote & pisco \\ 
    \midrule
    \texttt{pr} & decadron & scan & micoplasma & coqui & baclofen & mycoplasma \\ 
    \midrule
    \texttt{py} & paraguaya & itaipu & solano & plaqueta & paraguay & paraguayo \\ 
    \midrule
    \texttt{sv} & pital & parlacen & siguanaba & lempa & fulcro & salvador \\ 
    \midrule
    \texttt{us} & taxes & apocalypto & millions & contenedor & million & green \\ 
    \midrule
    \texttt{uy} & artigas & penarol & uruguaya & montevideo & sambayon & sapitos \\ 
    \midrule
    \texttt{ve} & camejo & hipolita & morrocoy & avanzadora & ribas & cardenales \\ 
    \midrule
    \bottomrule
\end{tabular}
    
  \caption{\textbf{Words in the queries of MessIRve are representative of the dialectal varieties of Spanish and the topics of interest in each country.}
    To find distinctive words for each country, we calculated the log odds ratio of the frequency of each word in each country with respect to the queries of the \texttt{full} subset, using smoothing with a prior of 0.1 \citep{jurafsky2024slp}. We consider only words with at least 10 occurrences in the country subset and show the top 6 words with the highest log odds ratio for each country.
  }
  \label{tab:spanish_varieties}
\end{table*}

\begin{table*}[ht!]
  \centering
  \small
  \begin{tabular}{|>{\centering\arraybackslash}m{3.5cm}|>{\centering\arraybackslash}m{3.5cm}|>{\centering\arraybackslash}m{3.5cm}|>{\centering\arraybackslash}m{3.5cm}|}
    \hline
    \textbf{MessIRve}        & \textbf{mMARCO}          & \textbf{MIRACL}          & \textbf{SQAC}          \\ \hline

    \textsc{Topic 2} & \textsc{Topic 5} & \textsc{Topic 19} & \textsc{Topic 0} \\
    n=61,797 (8.8\%) & n=60,499 (7.4\%) & n=498 (11.5\%) & n=4,314 (22.9\%) \\
    duele, razones, motivos, da, salen & costo, cuesta, salario, promedio, precio & paises, guerra, nazi, pais, imperio & cargo, profesion, equipo, partido, dirige \\
    \textit{porque da dolor y ardor en la boca del estomago} & \textit{costo promedio de la universidad} & \textit{¿Cuándo ocurrió la primera guerra mundial?} & \textit{¿En qué equipo juega Ronaldo?} \\
    \hline
    \textsc{Topic 3} & \textsc{Topic 1} & \textsc{Topic 0} & \textsc{Topic 19} \\
    n=48,188 (6.9\%) & n=59,825 (7.3\%) & n=455 (10.5\%) & n=2,693 (14.3\%) \\
    consiste, derechos, importante, sociales, cuales & significa, empresa, hace, archivo, red & cantante, banda, llama, personaje, pelicula & guerra, gobierno, pais, presidente, mundial \\
    \textit{porque son importante los derechos humanos} & \textit{la empresa de redes más rica del mundo} & \textit{¿Qué actor interpretó a Ray Charles en el cine?} & \textit{¿En qué continente tuvo lugar la Guerra de la Independencia Argentina?} \\
    \hline
    \textsc{Topic 4} & \textsc{Topic 6} & \textsc{Topic 25} & \textsc{Topic 4} \\
    n=45,908 (6.5\%) & n=47,105 (5.8\%) & n=380 (8.8\%) & n=1,733 (9.2\%) \\
    queda, region, pertenece, continente, rio & impuestos, credito, obtener, licencia, seguro & primer, primera, invento, fundo, creo & lugar, adonde, halla, encuentra, ciudad \\
    \textit{en qué continente queda australia} & \textit{¿Cuánto tiempo se tarda en obtener un reembolso de impuestos?} & \textit{¿Dónde se inventó el baloncesto?} & \textit{¿Con qué territorio limita Dubái por su parte occidental?} \\
    \hline
    \textsc{Topic 0} & \textsc{Topic 0} & \textsc{Topic 24} & \textsc{Topic 24} \\
    n=43,578 (6.2\%) & n=46,039 (5.6\%) & n=319 (7.4\%) & n=1,485 (7.9\%) \\
    murio, juega, canta, gano, equipo & murio, canta, elenco, interpreta, pelicula & biblia, historia, mitologia, religion, dios & obra, obras, segun, jesus, iglesia \\
    \textit{con quien canta michael jackson todo mi amor eres tu} & \textit{quien canta la cancion uno} & \textit{¿Quién era David en la Biblia?} & \textit{¿Quiénes escribieron los evangelios que se creen bastante cercanos a la época en que vivió Jesús de Nazaret?} \\
    \hline
    \textsc{Topic 11} & \textsc{Topic 9} & \textsc{Topic 4} & \textsc{Topic 11} \\
    n=43,539 (6.2\%) & n=44,004 (5.4\%) & n=313 (7.2\%) & n=1,139 (6.1\%) \\
    cuantas, cuantos, existen, municipios, jugadores & sintomas, enfermedad, causar, efectos, enfermedades & paises, capital, pais, ciudad, ciudades & personas, cuantas, cuantos, victimas, fallecieron \\
    \textit{cuántos idiomas indígenas existen en méxico} & \textit{¿Qué enfermedad causa dolor de estómago, diarrea y es contagiosa?} & \textit{¿En qué región se encuentra la República de las Islas Marshall?} & \textit{¿Cuántas personas utilizan este servicio en todo mundo?} \\
    \hline
    \textsc{Topic 7} & \textsc{Topic 8} & \textsc{Topic 13} & \textsc{Topic 20} \\
    n=36,811 (5.2\%) & n=43,946 (5.4\%) & n=312 (7.2\%) & n=978 (5.2\%) \\
    sirve, tomar, medicamento, mg, cura & numero, telefono, windows, computadora, excel & botanica, cuales, vegetacion, populares, caracteristicas & produjo, comicios, competicion, lugar, partido \\
    \textit{que pastilla puedo tomar para el dolor de cabeza} & \textit{Excel número de teléfono de servicio al cliente} & \textit{¿Cuántos colores puede reconocer un perro?} & \textit{¿Cómo se pasa del horario de invierno al de verano?} \\
    \hline
    \textsc{Topic 12} & \textsc{Topic 14} & \textsc{Topic 3} & \textsc{Topic 25} \\
    n=30,631 (4.4\%) & n=43,221 (5.3\%) & n=261 (6.0\%) & n=961 (5.1\%) \\
    energia, agua, tierra, ambiente, planeta & definicion, definir, significa, medica, biologia & principales, estudia, diferencia, cuales, paleontologos & primer, primera, escribio, creo, descubrio \\
    \textit{qué tipo de movimiento realiza el planeta tierra alrededor del sol} & \textit{definición de sobre} & \textit{¿Qué habilidades debe tener un director ejecutivo?} & \textit{¿Quién descubrió los rayos X?} \\
    \hline
    \textsc{Topic 1} & \textsc{Topic 23} & \textsc{Topic 15} & \textsc{Topic 1} \\
    n=28,177 (4.0\%) & n=38,566 (4.7\%) & n=240 (5.5\%) & n=636 (3.4\%) \\
    trata, consiste, marca, juego, gama & condado, encuentra, poblacion, tx, ca & botanica, significan, significa, idioma, sigla & siglas, dedica, drcafta, posicion, objetivo \\
    \textit{de que trata te para 3} & \textit{¿En qué condado se encuentra la ciudad de texas, tx?} & \textit{¿Qué quiere decir la palabra pseudo?} & \textit{¿Cómo se llama el programa?} \\
    \hline

\end{tabular}

  \caption{\textbf{Representative words and queries from topics in Spanish IR datasets.}
    These were selected based on the c-TF-IDF scores of the words and queries in each topic, respectively. Refer to Section \ref{sec:topic_analysis} for more details.
  }
  \label{tab:datasets_topics}
\end{table*}

\begin{table*}[ht!]
  \centering
  \small
  \begin{tabularx}{\textwidth}{%
  >{\centering\arraybackslash}m{0.07\textwidth} 
  >{\centering\arraybackslash}m{0.33\textwidth} 
  >{\scriptsize\raggedright\arraybackslash}m{0.52\textwidth}
}

\toprule
\textbf{Country} & \textbf{Query} & \textbf{Relevant document} \\
\midrule
\midrule
\texttt{ar} & quien era mama antula (\emph{who was Mama Antula?}) & María Antonia de Paz y Figueroa. Mama Antula se convirtió en la novena persona de nacionalidad argentina en ser beatificada. Esto ocurrió entre casi medio centenar de causas para canonizar [...] \\ 
\midrule
\texttt{bo} & a cuantos kilometros esta samaipata de santa cruz (\emph{how many kilometers is Samaipata from Santa Cruz?}) & Samaipata. La ciudad de Samaipata se encuentra a 119 km por carretera al suroeste de la capital departamental, Santa Cruz de la Sierra. Por Samaipata discurre la ruta troncal Ruta [...] \\ 
\midrule
\texttt{cl} & en qué región queda huilo huilo (\emph{in which region is Huilo Huilo located?}) & Reserva biológica Huilo Huilo. La Reserva Biológica Huilo Huilo es un área natural protegida privada que se ubica en medio de Los Andes, a 860 km al sur de Santiago [...] \\ 
\midrule
\texttt{co} & donde murio jorge eliecer gaitan (\emph{where did Jorge Eliécer Gaitán die?}) & Jorge Eliécer Gaitán. Un hombre (aparentemente Juan Roa Sierra u otros más) lo esperaba en la entrada del edificio y le disparó con un revólver causándole heridas mortales. Gaitán fue [...] \\ 
\midrule
\texttt{cr} & que visitar en puntarenas (\emph{what to visit in Puntarenas?}) & Puntarenas (ciudad). La ciudad de Puntarenas, como capital de la provincia, cuenta con diversos lugares históricos de interés turístico, entre los que se pueden mencionar la antigua Capitanía del Puerto [...] \\ 
\midrule
\texttt{cu} & que hizo antonio maceo por cuba (\emph{what did Antonio Maceo do for Cuba?}) & Antonio Maceo. José Antonio de la Caridad Maceo y Grajales (San Luis, Santiago de Cuba, 14 de junio de 1845 - San Pedro, La Habana, 7 de diciembre de 1896) [...] \\ 
\midrule
\texttt{do} & qué hizo enriquillo (\emph{what did Enriquillo do?}) & Enriquillo. Enrique Bejo (lago Jaragua, Cacicazgo de Jaragua, ca. 1498-Sabana Buey, Llano de Baní, 27 de septiembre de 1535), más conocido como Enriquillo, fue un noble taíno que se alzó [...] \\ 
\midrule
\texttt{ec} & de que region son los montubios (\emph{from which region are the Montubios?}) & Montuvio. Según el Censo ecuatoriano de 2022 los montuvios representan el 7,7\% de la población del Ecuador, lo que significa que en el 2022 más de 1.304.994 ecuatorianos se identificaron [...] \\ 
\midrule
\texttt{es} & de qué se pueden hacer las croquetas (\emph{what can croquettes be made from?}) & Croqueta. A veces, pero no siempre, se enharinan primero y se sacude la harina sobrante. Se bañan en huevo batido y se cubren siempre generosamente con pan rallado. Después se [...] \\ 
\midrule
\texttt{gt} & dónde queda mazatenango (\emph{where is Mazatenango?}) & Mazatenango. Mazatenango (del náhuatl, significa «"muralla del venado"») es una ciudad y cabecera del departamento de Suchitepéquez, en la República de Guatemala. El municipio se encuentra localizado a 161 km [...] \\ 
\midrule
\texttt{hn} & quien y en que año se fundo la villa de san pedro sula (\emph{who and in which year was the town of San Pedro Sula founded?}) & San Pedro Sula. San Pedro Sula fue fundada el 27 de junio de 1536, bajo el nombre de San Pedro de Puerto Caballos, por el conquistador español Pedro de Alvarado. [...] \\ 
\midrule
\texttt{mx} & maluma carin leon - según quién versuri romana (\emph{Maluma Carin León - according to whom, Roman lyrics?}) & Según quién. «Según quién» es una canción del cantante colombiano Maluma y el cantante mexicano Carín León. Fue lanzada el 17 de agosto de 2023, a través de Sony Music [...] \\ 
\midrule
\texttt{ni} & cuándo hizo erupción el volcán masaya (\emph{when did Masaya Volcano erupt?}) & Volcán Masaya. Otras erupciones han ocurrido en los últimos 50 años. El 22 de noviembre de 1999 empezó un nuevo evento eruptivo, apareciendo un punto rojo en las imágenes de [...] \\ 
\midrule
\texttt{pa} & que año nacio delfia cortez (\emph{in what year was Delfia Cortez born?}) & Delfia Cortez. Delfia Nereida Cortez Marciaga (Los Santos, 21 de enero de 1963-Ciudad de Panamá, 10 de enero de 2024), fue una periodista y política panameña. \\ 
\midrule
\texttt{pe} & quién es dina boluarte (\emph{who is Dina Boluarte?}) & Dina Boluarte. Fue electa primera vicepresidenta en las elecciones presidenciales de Perú de 2021, y ocupó el cargo desde el 28 de julio de 2021 hasta el 7 de diciembre [...] \\ 
\midrule
\texttt{pr} & que hace el decadron inyectable (\emph{what does injectable Decadron do?}) & Dexametasona. La dexametasona es un potente glucocorticoide sintético con acciones que se asemejan a las de las hormonas esteroides. Actúa como antiinflamatorio e inmunosupresor. Su potencia de 20-30 veces la [...] \\ 
\midrule
\texttt{py} & cómo surgió la nación paraguaya (\emph{how did the Paraguayan nation arise?}) & Historia de Paraguay. Lo que actualmente es el territorio de Paraguay fue descubierto por Alejo García y Juan de Ayolas, a las órdenes de España en 1524, dándose inicio a [...] \\ 
\bottomrule

\end{tabularx}

  \caption{\textbf{Example query-document pairs per country.}
    Queries are sampled from those containing the top log-odds ratio words per country, as listed in Table \ref{tab:spanish_varieties}. 
    For readability, queries were translated into English by the author, and only the first 30 words of each document are displayed.
  }
  \label{tab:dataset_examples}
\end{table*}

\end{document}